# Temporal Output Discrepancy for Loss Estimation-Based Active Learning

Siyu Huang, Tianyang Wang, Haoyi Xiong, *Senior Member, IEEE*, Bihan Wen, *Member, IEEE*, Jun Huan, *Senior Member, IEEE*, and Dejing Dou, *Senior Member, IEEE*

*Abstract*—While deep learning succeeds in a wide range of tasks, it highly depends on the massive collection of annotated data which is expensive and time-consuming. To lower the cost of data annotation, active learning has been proposed to interactively query an oracle to annotate a small proportion of informative samples in an unlabeled dataset. Inspired by the fact that the samples with higher loss are usually more informative to the model than the samples with lower loss, in this article we present a novel deep active learning approach that queries the oracle for data annotation when the unlabeled sample is believed to incorporate high loss. The core of our approach is a measurement temporal output discrepancy (TOD) that estimates the sample loss by evaluating the discrepancy of outputs given by models at different optimization steps. Our theoretical investigation shows that TOD lower-bounds the accumulated sample loss thus it can be used to select informative unlabeled samples. On basis of TOD, we further develop an effective unlabeled data sampling strategy as well as an unsupervised learning criterion for active learning. Due to the simplicity of TOD, our methods are efficient, flexible, and task-agnostic. Extensive experimental results demonstrate that our approach achieves superior performances than the state-of-the-art active learning methods on image classification and semantic segmentation tasks. In addition, we show that TOD can be utilized to select the best model of potentially the highest testing accuracy from a pool of candidate models.

## I. Introduction

LARGE-SCALE annotated datasets are indispensable and critical to the success of modern deep learning models. Since the annotated data are often highly expensive to obtain, learning techniques including unsupervised learning [1], semi-supervised learning [2], and weakly supervised learning [3] have been widely explored to alleviate the dilemma. In this article, we focus on active learning [4] which aims to selectively annotate unlabeled data with limited budgets while resulting in high-performance models.

In existing literature of active learning, two mainstream approaches have been studied, namely the diversity-aware approach and the uncertainty-aware approach. The diversity-aware approach [5] aims to pick out diverse samples to represent the distribution of a dataset. It works well on low-dimensional data and classifier with a small number of classes [6]. The uncertainty-aware approach [7], [8] aims to pick out the most uncertain samples based on the current model. However, the uncertainty heuristics, such as distance to decision boundary [9] and entropy of posterior probabilities [10], are often task-specific and need to be specifically designed for individual tasks such as image classification [11], object detection [12], and semantic segmentation [13].

In this article, we consider that the samples with higher loss would be more informative than the ones with lower loss. Specifically in supervised learning settings, when samples are correctly labeled, the averaged loss function over all samples should be gradually minimized during the learning procedure. Moreover, in every iteration the training model would backward propagated error according to the loss of every sample [14], while the sample with high loss usually brings informative updates to the parameters of the training model [15]. In this work, we extend these evidences to active learning problems and propose a simple yet effective loss estimator temporal output discrepancy (TOD), which can measure the potential loss of a sample only relied on the training model, when the ground-truth label of the sample is not available. Specifically, TOD computes the discrepancy of outputs given by models at different optimization steps, and a higher discrepancy corresponds to a higher sample loss. Our theoretical investigation shows that TOD well measures the sample loss.

On basis of TOD, we propose a deep active learning framework that leverages a novel unlabeled data sampling strategy for data annotation in conjunction with a semisupervised training scheme to boost the task model performance with unlabeled data. Specifically, the active learning procedure can be split into a sequence of training cycles starting with a small number of labeled samples. By the end of every training cycle, our data sampling strategy estimates cyclic

Manuscript received 27 October 2021; revised 11 February 2022 and 3 May 2022; accepted 19 June 2022. This work was supported in part by the NIH under Grant 5U54CA225088-03 and in part by the Ministry of Education, Republic of Singapore, under its Academic Research Fund Tier 1 under Project RG137/20 and Start-up Grant. *(Corresponding author: Bihan Wen.)*
Siyu Huang is with the Harvard John A. Paulson School of Engineering and Applied Sciences, Harvard University, Cambridge, MA 02134 USA (e-mail: huang@seas.harvard.edu).
Tianyang Wang is with the Department of Computer Science and Information Technology, Austin Peay State University, Clarksville, TN 37044 USA (e-mail: wangt@apsu.edu).
Haoyi Xiong and Dejing Dou are with the Big Data Laboratory, Baidu Research, Beijing 100193, China (e-mail: xionghaoyi@baidu.com; doudejing@baidu.com).
Bihan Wen is with the School of Electrical and Electronic Engineering, Nanyang Technological University, Singapore 639798 (e-mail: bihan.wen@ntu.edu.sg).
Jun Huan is with the AWS AI Lab, Amazon, Seattle, WA 98109 USA (e-mail: lukehuan@amazon.com).
Color versions of one or more figures in this article are available at https://doi.org/10.1109/TNNLS.2022.3186855.
Digital Object Identifier 10.1109/TNNLS.2022.3186855

output discrepancy (COD), which is a variant of TOD, for every sample in the unlabeled pool and selects the unlabeled samples with the largest COD for data annotation. The newly annotated samples are added to the labeled pool for model training in the next cycles. The task learning objective is augmented with a regularization term derived from TOD, so as to improve the performance of active learning with the aid of the unlabeled samples. Compared with the existing deep active learning algorithms, our approach is more efficient, more flexible, and easier to implement, since it does not introduce extra learnable models such as the loss prediction module [16] or the adversarial network [17], [18] for uncertainty estimation.

In addition to active learning task, we also study the model selection task, where the goal is to select the models of superior performances from a pool of well-trained models. Although it is a widely adopted manner that the trained models are evaluated on the validation set then the models of the best performance are picked out, there would be a nontrivial domain gap between the validation set and test set. Directly evaluating models on the unlabeled test set[1] would be an effective solution to alleviate this gap when the unlabeled test set is accessible. In this work, we show that the proposed loss measure TOD can also be used to pick out the models of potentially high testing accuracy.

In the experiments, our active learning approach shows superior performances in comparison with the state-of-the-art baselines on various image classification and semantic segmentation datasets. Extensive ablation studies demonstrate that our proposed TOD can well estimate the sample loss and benefit both the active data sampling and the task model learning.

The contributions of this article are summarized as follows.

1) This article proposes a simple yet effective loss measure TOD. Both theoretical and empirical studies validate the efficacy of TOD.
2) This article presents a novel deep active learning method, which includes a TOD-based active sampling strategy and a semisupervised learning scheme.
3) This article further presents a TOD-based model selection algorithm to address both model-level and sample-level model selection tasks.
4) Extensive active learning experiments on image classification and semantic segmentation tasks evaluate the effectiveness of the proposed methods.

## II. Related Work

### A. Active Learning

Active learning aims to incrementally annotate samples that result in high model performance and low annotation cost [4], [19]–[21]. Active learning has been studied for decades of years and the existing methods can be generally grouped into two categories: the query-synthesizing approach and the query-acquiring approach. The query-synthesizing approach [22], [23] employs generative models to synthesize new informative samples. For instance, adversarial sampling for active learning (ASAL) [24] uses generative adversarial networks (GANs) [25] to generate high-entropy samples. In this article, we focus on the query-acquiring active learning which develops effective data sampling strategies to pick out the most informative samples from the unlabeled data pool.

The query-acquiring methods can be categorized as diversity-aware and uncertainty-aware methods. The diversity-aware methods [5], [26] select a set of diverse samples that best represents the dataset distribution. A typical diversity-aware method is the core-set selection [6] based on the core-set distance of intermediate features. It is theoretically and empirically proven to work well with a small scale of classes and data dimensions.

The uncertainty-aware methods [7], [27]–[30] actively select the most uncertain samples in the context of the training model. A wide variety of related methods has been proposed, such as Monte Carlo estimation of expected error reduction [31], distance to the decision boundary [9], [32], margin between posterior probabilities [33], [34], and entropy of posterior probabilities [10], [11], [35].

The diversity-aware and uncertainty-aware approaches are complementary to each other. For instance, the uncertainty-aware active learning methods do not take sample redundancy into consideration, such that similar high-uncertainty samples would be picked out in an active learning cycle. Therefore, many hybrid methods [36]–[45] have been proposed for specific tasks. In more recent literature, adversarial active learning [17], [18], [46] is introduced to learn an adversarial discriminator to distinguish the labeled and unlabeled data.

Compared to the existing works in active learning, our method falls into the category of uncertainty-aware active learning by directly utilizing the task model for uncertainty estimation. The relevant works include the ones which utilize the expected gradient length [47] or output changes on input perturbation [48], [49] for uncertainty estimation. In the realm of loss estimation, Yoo and Kweon [16] proposed to learn a loss prediction module to estimate the loss of unlabeled samples. However, different from [16], our method only relies on the task model without learning extra models such as the loss prediction module [16] or the adversarial network [17], [18]. Our method is very efficient since its computation only includes the feed-forward inference of the task model. In addition, we provide a theoretical interpretation for our method by connecting it to the lower bound of the accumulated sample loss.

### B. Semisupervised Learning

This work is also related to semisupervised learning which seeks to learn from both labeled and unlabeled data since we also develop the proposed loss estimation method to improve the learning of task model using unlabeled data. There has been a wide variety of semisupervised learning approaches such as transductive model [50], graph-based method [51], and generative model [52]. We refer to [53] for an up-to-date overview.

More recently, several semisupervised methods including Π-model [54] and virtual adversarial training [55]

---

[1]The unlabeled test set is visible to the machine learning algorithm developers under some circumstances, e.g., Kaggle competitions.

apply consistency regularization to the posterior distributions of perturbed inputs. Further improvements including Mean Teacher [56] and Temporal Ensembling [54] apply the consistency regularization on models at different time steps. However, the consistency regularization has been seldom exploited for active learning.

Compared to the existing efforts in semisupervised learning for neural networks, our proposed loss measure TOD could be considered as an alternative solution of consistency regularization. TOD can be well adapted to active learning by developing a novel active sampling method COD. COD only relies on the models learned after every active learning cycle. In contrast, the existing temporal consistency-based uncertainty measurements often require access to a number of previous model states. For instance, the computing of Mean Teacher [56] and Temporal Ensembling [54] require the historical model parameters and the historical model outputs, respectively.

On the other hand, there have not been sufficient theoretical interpretations for the success of consistency regularization. Athiwaratkun et al. [57] reveals that the consistency regularization on perturbed inputs is an unbiased estimator for the norm of the Jacobian of the network. However, there is still a lack of interpretations on the temporal consistency regularization. In this article, we show that the temporal consistency regularization can be connected to the lower bound of the accumulated sample loss. Thus, the temporal consistency regularization is a theoretically effective solution to loss estimation as well as semisupervised learning.

## III. Temporal Output Discrepancy

Measuring the sample loss on a given neural network $f$, when the label of the sample is unavailable, is a key challenge for many learning problems, including active learning [29], [33], [35], continual learning [30], and self-supervised learning [54], [56]. In this work, we present TOD, which estimates the sample loss based on the discrepancy of outputs of a neural network at different learning iterations. Given a sample $x \in \mathbb{R}^d$, we have TOD $D_t^{\{T\}} : \mathbb{R}^d \to \mathbb{R}$

$$D_t^{\{T\}}(x) \stackrel{\text{def}}{=} \|f(x; w_{t+T}) - f(x; w_t)\|. \tag{1}$$

$D_t^{\{T\}}(x)$ characterizes the distance[2] between outputs of model $f$ with parameters $w_{t+T}$ and $w_t$ obtained in the $(t+T)$th and $t$th gradient descend step during learning (e.g., $T > 0$), respectively.

In the following, we show that a larger $D_t^{\{T\}}(x)$ indicates a larger sample loss[3] $L_t(x) = (1/2)(y - f(x; w_t))^2$, where $y \in \mathcal{R}$ is the label corresponding to sample $x$. We first give the upper bound of one-step output discrepancy $D_t^{\{1\}}(x)$.

*Theorem 1:* With an appropriate setting of learning rate $\eta$

$$D_t^{\{1\}}(x) \leq \eta \sqrt{2L_t(x)} \|\nabla_w f(x; w_t)\|^2. \tag{2}$$

[2]For brevity, $\|\cdot\|$ denotes the $L_2$ norm $\|\cdot\|_2$ in this article.
[3]Here we take Euclidean loss as an example. The cross-entropy (CE) loss has similar results.

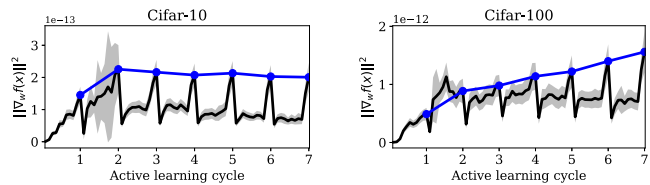

Fig. 1. $\|\nabla_w f\|^2$ versus the active learning cycle. The dark lines denote $\|\nabla_w f\|^2$ averaged over the training (i.e., labeled and unlabeled) pool. The blue lines denote average $\|\nabla_w f\|^2$ after every active learning cycle.

The proofs of Theorem 1 and the following corollaries can be found in the Appendix. From Theorem 1, the upper bound of $T$-step output discrepancy $D_t^{\{T\}}(x)$ can be easily deduced.

*Corollary 1:* With an appropriate setting of learning rate $\eta$

$$D_t^{\{T\}}(x) \leq \sqrt{2}\eta \sum_{\tau=t}^{t+T-1} \left(\sqrt{L_\tau(x)} \|\nabla_w f(x; w_\tau)\|^2\right). \tag{3}$$

Corollary 3 preliminarily connects $T$-step output discrepancy $D_t^{\{T\}}(x)$ to sample loss $L(x)$. However, it is almost infeasible to compute $\|\nabla_w f(x; w_\tau)\|$ on all $\tau$. Fortunately, $\|\nabla_w f\|$ is approximately a constant under the context of neural networks, as discussed in [59] and [60].

*Remark 1:* For a linear layer $\phi(x; W)$ with rectified linear unit (ReLU) activation, the Lipschitz constant $\mathcal{L}(W) \leq \|x\|$.

Since sample $x$ is drawn from a distribution $X$, we assume $\|x\|$ is a constant so that $f$ is Lipschitz-continuous over $w$. Thus, we let $\|\nabla_w f\|^2$ be upper-bounded by a constant $C$. Empirical results on image classification benchmarks including Cifar-10 and Cifar-100 also support this assumption. As shown in Fig. 1, the dark lines are the averaged $\|\nabla_w f\|^2$ over the training set. The blue lines denote the averaged $\|\nabla_w f\|^2$ after every active learning cycle. $\|\nabla_w f\|^2$ has a small variance over samples and it is nearly constant across every active learning cycle.

With $\|\nabla_w f\|^2 \leq C$, we rewrite Corollary 3 to connect $D_t^{\{T\}}(x)$ with the accumulated loss of sample $x$.

*Corollary 2:* With appropriate settings of a learning rate $\eta$ and a constant $C$

$$D_t^{\{T\}}(x) \leq \sqrt{2T}\eta C \sqrt{\sum_{\tau=t}^{t+T-1} L_\tau(x)}. \tag{4}$$

Corollary 4 shows that $\|f(x; w_{t+T}) - f(x; w_t)\|$ is a lower bound of the square root of accumulated loss $L$ during $T$ gradient descend steps. Thus, when $T$ is fixed, e.g., a certain number of iterations of neural network training, TOD can effectively estimate the loss of sample $x$. Note that the preassumptions of Theorem 1 and its corollaries limit the learning rate $\eta$ not to be too large to dissatisfy the Taylor expansion used in our proofs. In empirical study, we find that the commonly used learning rates, e.g., $\eta = 0.1$ or smaller, work well.

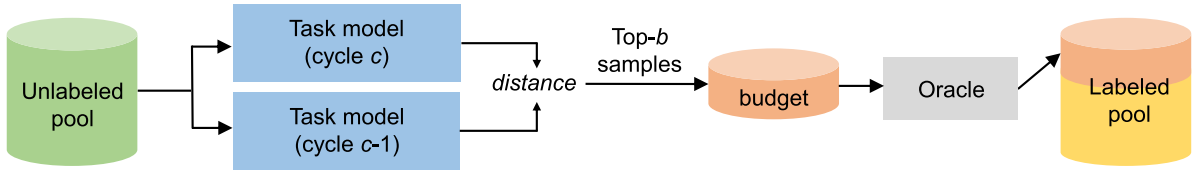

Fig. 2. COD-based unlabeled data sampling strategy for active learning. Data samples with the largest COD are collected from the unlabeled pool. The collected samples are annotated by an oracle and added to the labeled pool.

## IV. SEMISUPERVISED ACTIVE LEARNING

### A. Problem Formulation

We first formulate the standard active learning task as follows. Let $(x_S, y_S)$ denote a sample pair drawn from a set of labeled data $(X_S, Y_S)$, where $X_S$ is the data points and $Y_S$ is the labels. Let $x_U$ denote an unlabeled sample drawn from a larger unlabeled data pool $X_U$, i.e., the labels $Y_U$ corresponding to $X_U$ cannot be observed. In an active learning cycle $c$, the active learning algorithm selects a fixed budget of samples from the unlabeled pool $X_U$ and the selected samples will be annotated by an oracle. The budget size $b$ is usually much smaller than $|X_U|$, the size of the unlabeled pool. The goal of active learning is to select the most informative unlabeled samples for annotation, so as to minimize the expected loss of a task model $f : X \to Y$.

We next present the use of TOD in a *semisupervised* active learning framework. An active learning algorithm generally consists of two components: 1) an unlabeled data sampling strategy and 2) the learning of a task model. We adapt TOD to these two components, respectively. For component 1), we propose COD, a new criterion to select unlabeled samples with the largest *estimated loss* for annotation. For component 2), we develop a TOD-based unsupervised loss term to improve the performance of task model. In the following, we formulate the active learning problem and discuss the details of the two components.

### B. Cyclic Output Discrepancy

In (4), our proposed TOD characterizes a lower bound of the loss function for supervised learning. Here, we introduce a variant of TOD, i.e., COD, for active selection of unlabeled samples. COD estimates the sample uncertainty by measuring the difference of model outputs between two consecutive active learning cycles

$$D_{\text{cyclic}}(x|w_c, w_{c-1}) = \|f(x; w_c) - f(x; w_{c-1})\| \quad (5)$$

where model parameters $w_c$ and $w_{c-1}$ are obtained after the $c$th and $(c-1)$th active learning cycle, respectively.

Fig. 2 illustrates the procedure of COD-based unlabeled data sampling. Given COD for every sample in unlabeled pool $X_U$, our strategy selects $b$ samples with the largest COD from $X_U$. Then, the strategy queries human oracles for annotating the selected samples. The newly annotated data is added to labeled pool for the next active learning cycle. In the first active learning cycle (i.e., $c = 1$), the model $f$ is trained with a random set of labeled data, then COD is computed based on the initial model and the model learned after the first cycle. For $c \geq 2$, we compute COD $D_{\text{cyclic}}(x|w_c, w_{c-1})$ for active sample selection systematically.

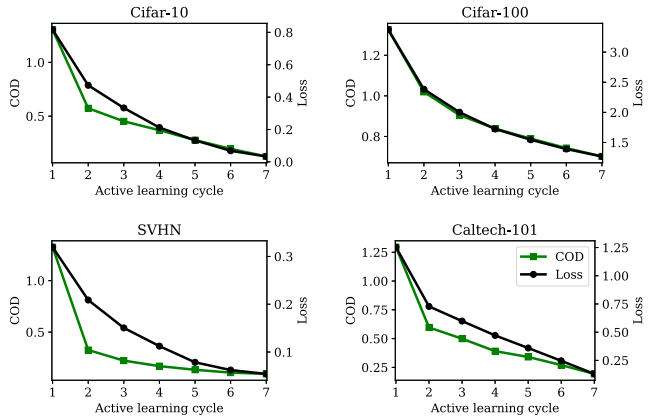

Fig. 3. Consistency of COD and the real task loss. We show COD and real loss averaged over the unlabeled samples versus percentage of labeled images, under active learning setting.

*Minimax Optimization of COD:* As discussed in Corollary 4, COD-based data sampling strategy can find samples of large loss in unlabeled pool, so as to minimize the expected loss of model $f$ through further training the task model in the next cycle. Fig. 3 preliminarily verifies the consistency between COD and the real loss, where COD shows a similar trend with the real loss and they are both decreasing along with the active learning progress. Instead of minimizing the TOD directly (i.e., min-min optimization which may not be good), COD develops TOD as the criterion of sample selection in active learning, where samples with the maximal TOD are picked up (e.g., max-min strategies). When considering labels of samples with potential losses as information gain, our strategy actually maximizes the minimum gain in active learning.

### C. Semisupervised Task Learning

As suggested by Corollary 4, TOD measures the accumulated sample loss, and thus it is natural to employ TOD as an unsupervised criterion to improve the learning of model $f$ using the unlabeled data. However, directly applying TOD to unsupervised training with the baseline model obtained at the last cycle $c - 1$ may lead to an unstable training, due to the following aspects: 1) the iteration interval between current model and baseline model (i.e., $T$ in Corollary 4) is no longer fixed during model training, thus the loss measurement would be inaccurate and 2) the baseline model only depends on a single historical model state so that it may suffer from a large variance in loss measurement.

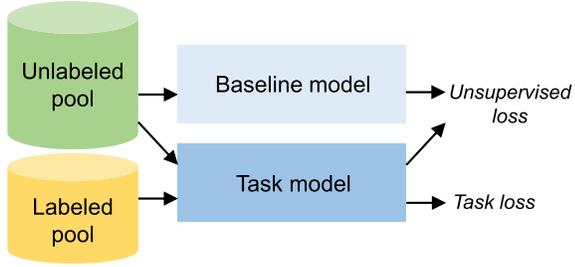

Fig. 4. Semisupervised task learning scheme. For labeled data, the task model is trained with the task loss. For unlabeled data, the task model is trained to minimize the distance between outputs of the task model and the baseline model.

To address the above issues, we exploit a Mean Teacher [56] to memorize model parameters over the long-term training procedure. The parameters $\tilde{w}$ of the baseline model are the exponential moving average (EMA) of the historical parameters, as

$$\tilde{w} \leftarrow \alpha \cdot \tilde{w} + (1 - \alpha) \cdot w \qquad (6)$$

where $w$ is the parameters of the current model and $\alpha$ is the EMA decay rate. We propose an unsupervised criterion that minimizes the L2 loss between the current model and the baseline model. In the $c$th cycle, with the unlabeled pool $X_U^c$, the unsupervised criterion is

$$L_U^c(w) = \frac{1}{|X_U^c|} \sum_{x_U \in X_U^c} \|f(x_U; w) - f(x_U; \tilde{w})\|^2. \qquad (7)$$

We next present the semisupervised task learning scheme. As illustrated in Fig. 4, for labeled data, we optimize the supervised task objective. Here, we take the CE loss for image classification as an example. In the $c$th cycle, given the labeled set $(X_S^c, Y_S^c)$ in the cycle, the supervised loss is

$$L_S^c(w) = \frac{1}{|X_S^c|} \sum_{(x_S, y_S) \in X_S^c, Y_S^c} CE[f(x_S; w), y_S]. \qquad (8)$$

Note that the labeled pool $(X_S^c, Y_S^c)$ will be enlarged per active learning cycle. Within an active learning cycle, the labeled pool remains unchanged.

By integrating the supervised task objective and the unsupervised loss, we formulate the overall loss function that evolves with the cycle $c$, as

$$L_{\text{overall}}^c(w) = L_S^c(w) + \lambda \cdot L_U^c(w) \qquad (9)$$

where $\lambda$ is a trade-off weight to balance the supervised and unsupervised loss terms. In our experiments, $\lambda$ is set to 0.05 and the EMA decay rate $\alpha$ is set to 0.999. See the Appendix for more details.

## V. TEST DATA-AWARE MODEL SELECTION

### A. Problem Formulation

In addition to active learning which selects samples of high losses from the unlabeled pool, we further introduce another application of TOD, i.e., test data-aware model selection, which aims at selecting models of potentially high testing accuracy from a pool of candidate models.

In general, the machine learning algorithms split the data into three parts: the *training set*, the *validation set*, and the *test set*. The *training set* is used to train the machine learning models. The *validation set* is used to select the algorithms, models, and hyperparameters after model training. The *test set* is unseen during training, and it is used to evaluate the final model performance. However, this model selection paradigm may fail when there is a nontrivial gap between the *training/validation sets* and the *test set* [61], [62]. Directly evaluating the models on the unlabeled *test set* becomes an effective alternative solution when *test set* is available. The uncertainty-aware [10], [33], [35] and the loss estimation-aware (e.g., LL4AL [16] and TOD) active learning methods are able to estimate the uncertainty/loss of the test data, which is a surrogate criterion of the testing accuracy. Thus, these methods can be utilized for test data-aware model selection. The problem can be formulated as follows. Given a neural network model $f$ with several independently trained parameters $w^{(1)}, w^{(2)}, \ldots, w^{(N)}$, the goal is to estimate the relative performance of these parameters on the unlabeled test set $X_{\text{test}}$.

### B. Method

According to Corollary 4, TOD is a lower bound of the accumulated sample loss $L$. Thus, the task of test accuracy prediction is relaxed to the estimation of its lower bound, i.e., the average TOD of $f(w^{(i)})$ on test set $X_{\text{test}}$. We have

$$D^{(i)} \leftarrow \frac{1}{|X_{\text{test}}|} \sum_{x \in X_{\text{test}}} \|f(x; w^{(i)}) - f(x; \tilde{w}^{(i)})\|^2 \qquad (10)$$

where $\tilde{w}^{(i)}$ is the baseline parameter corresponding to the trained parameter $w^{(i)}$. It is set as a certain previous parameter state during model training. In practice, we find that one optimization epoch between $\tilde{w}^{(i)}$ and $w^{(i)}$ works well for model selection. More empirical studies on the optimization iteration intervals can be found in Section VI-E. Although our algorithms needs access to the previous model checkpoint, we believe it might not be a heavy burden for users who prepare to use our algorithm to store a previous model checkpoint during training.

We illustrate the TOD-based model selection method in Algorithm 1. For each parameter $w^{(i)}$ of the candidate parameter pool, we compute the average TOD $D^{(i)}$ over the test set $X_{\text{test}}$. A smaller $D^{(i)}$ indicates a smaller test loss, i.e., a higher test accuracy. Finally, the sorting index of $D^{(i)}$ is a ranking list of the parameters $w^{(i)}$ which can be used for model selection.

**Algorithm 1** TOD-Based Model Selection Method

**Input:** unlabeled test set $X_{\text{test}}$
neural network model $f$
model weights $w^{(1)}, w^{(2)}, \ldots, w^{(N)}$
baseline weights $\tilde{w}^{(1)}, \tilde{w}^{(2)}, \ldots, \tilde{w}^{(N)}$
**Output:** model ranking list $r$
1: **for** $i = 1, 2, \ldots, N$ **do**
2: $\quad D^{(i)} \leftarrow \frac{1}{|X_{\text{test}}|} \sum_{x \in X_{\text{test}}} \|f(x; w^{(i)}) - f(x; \tilde{w}^{(i)})\|^2$;
3: **end for**
4: index $r \leftarrow$ sort $D$ in an ascending order;
5: **return** $r$

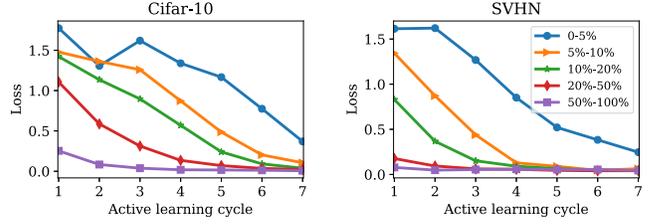

Fig. 5. Average real losses of unlabeled samples in a descending order of COD values. For instance, "0%–5%" denotes the 5% unlabeled samples which have the largest COD values, and so on.

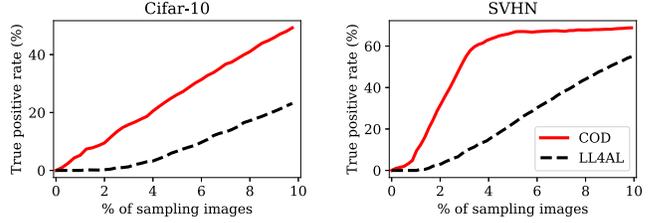

Fig. 6. Performance of loss estimation using a learned loss prediction model (LL4AL) [16] and the proposed COD method. We show the proportion of sampled images which have the highest real losses versus the proportion of sampling images.

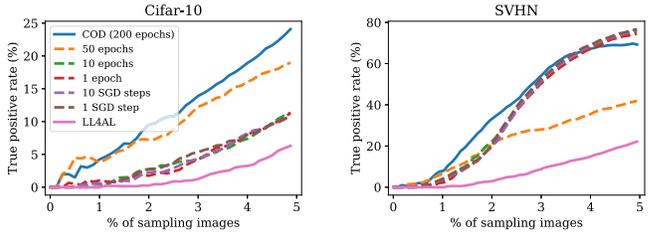

Fig. 7. Effects of number of GD steps used in COD.

## VI. EXPERIMENTS

We conduct extensive experimental studies to evaluate the proposed active learning approach on two computer vision tasks, image classification, and semantic segmentation, using seven benchmark datasets. The results are reported over three runs with different initial network weights and labeled pools. We implement the methods using PyTorch framework [63]. See the Appendix for more details.

### A. Efficacy of TOD as Loss Measure

This work proposes TOD to estimate the loss of an unlabeled sample. Fig. 3 has evaluated the relations between TOD and sample loss as discussed in Theorem 1 and Corollary 4, suggesting that the average COD and average loss have a consistent trend along with active learning cycles. To further verify the effectiveness of TOD for loss estimation, we study the average loss of unlabeled samples by sorting their COD values. Fig. 5 shows that the larger COD values of samples indicate the higher losses of samples, and, this observation is consistent across all the active learning cycles.

In Fig. 6, we compare the loss estimation performance of a learned loss prediction model (LL4AL) [16] and COD. We investigate how many samples of the highest losses can be picked out by using different methods. Fig. 6 shows that COD performs significantly better than LL4AL, as COD is able to pick out more high-loss samples under all the sampling settings. Figs. 3, 5, and 6 demonstrate that COD is an effective loss measure as well as a feasible criterion for active data sampling.

Fig. 7 shows the loss estimation performances of COD using different number of GD steps. Regarding the number of GD steps in COD, we conjecture there is a tradeoff between variance and bias: fewer GD steps indicate a more unbiased loss estimation with respect to current model, while more GD steps result in a smaller estimation variance. Fig. 7 shows that more GD steps generally lead to a better loss estimation performance especially when there are fewer sampling images. For SVHN dataset, 200-epoch COD and 50-epoch COD perform worse than the other methods when there are more sampling images. One possible cause is that the task model of SVHN has a more severe overfitting issue compared to those of other datasets, such that more GD steps may not significantly reduce the estimation variance, but introducing more biases into estimation to hurt the performance. To achieve a consistency between datasets, in this article, we adopt 200 training epochs, i.e., one active cycle, as the gap of COD.

### B. Active Learning for Image Classification

*1) Experimental Setup:* We evaluate active learning methods on four benchmark image classification datasets including Cifar-10 [64], Cifar-100 [64], Fashion-Mnist [65], SVHN [66], Caltech-101 [67], and STL-10 [68]. Following the conventional practices in deep active learning [16], [17], we employ ResNet-18 [69] as the image classification model. We compare our active learning approach against the state-of-the-art methods including Core/Uncertain GCN [70], SRAAL [18], TA-VAAl [71], VAAL [17], LL4AL [16], Core-set [6], and MC-Dropout [28]. In addition, the random selection of unlabeled data ("Random") and the model trained on the full training set ("Full Training") are also included as baselines.

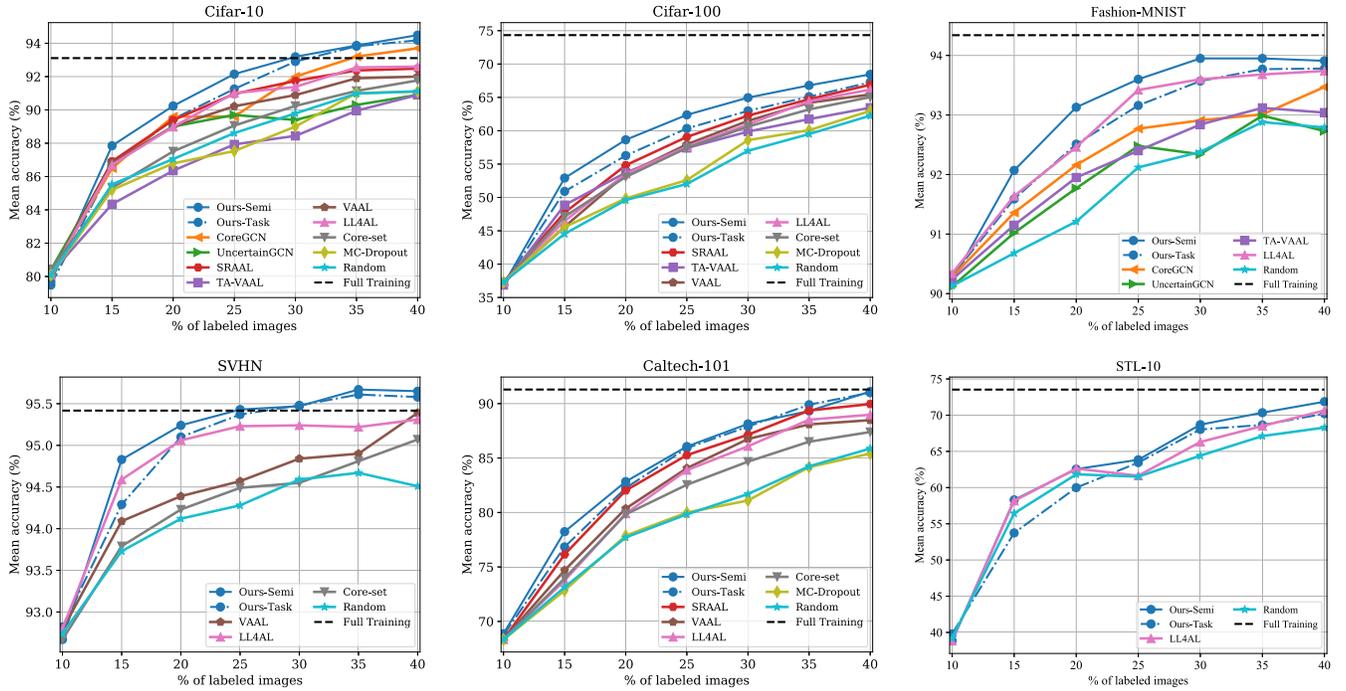

Fig. 8. Active learning results of image classification on six benchmark datasets.

*2) Results:* Fig. 8 shows image classification performances of different active learning methods. Our method outperforms all the other methods on the benchmark datasets. Additionally, we have the following observations.

1) Our method consistently performs better than the other methods with respect to the cycles. This is a desired property for a successful active learning method, since the labeling budget may vary for different tasks in real-world applications. For instance, one may only be able to annotate 20% instead of 40% of all the data.
2) Our method shows robust performances on difficult datasets such as Cifar-100 and Caltech-101. Both datasets include much more classes than Cifar-10, and Caltech-101 includes images of much higher resolution (i.e., 300 × 200). These difficult datasets bring more challenges to active learning, and the superior performances on these datasets demonstrate the robustness of our method.
3) On Cifar-10 and SVHN datasets which have many redundant data samples, the proposed method shows promising results compared to the diversity-aware methods including CoreGCN [70] and Core-Set [6] although it does not consider the diversity of selected samples.
4) The performance curves of our method are relatively smooth compared with the other methods. A smooth curve means there are consistent performance improvements from cycle to cycle, indicating that our sampling strategy can take informative data from the unlabeled pool.
5) In later cycles, our method shows a relatively larger accuracy improvement. We conjecture this is due to the utilization of unlabeled data in model training.
6) Our method uses 40% training samples to outperform the full training on Cifar-10 and SVHN, e.g., 94.5% versus 93.1% on Cifar-10. This interesting finding is in accord with the observations discussed in previous literature [72] that some data in the original dataset might be unnecessary or harmful to model training.

Table I compares different active learning methods, i.e., the state-of-the-art algorithms and the proposed methods, for image classification on 40% training labeled data. Both semisupervised task learning and active data selection strategy contributes to performance improvement, while, active data selection results in a more significant improvement than semisupervised task learning. We also note that the proposed method can outperform existing algorithms without semisupervised task learning (see "Base + Active" in Table I).

To study the robustness of active learning methods against label noise, we further conduct noisy active learning experiments on Cifar-10 and Cifar-100. We randomly perturb the groundtruth labels of training set with a probability of 20%. The results are shown in Fig. 9. Our COD outperforms LL4AL and random baseline on most of the learning cycles, indicating that COD enables a better model robustness to noisy labeling. In Fig. 8, several methods approach or even beat full training on standard active learning. However, all the methods cannot outperform full training on noisy active learning. It indicates

TABLE I
ACTIVE LEARNING PERFORMANCES ON 40% LABELED DATA. "BASE": STANDARD TASK MODEL TRAINING WITHOUT ACTIVE DATA SELECTION.
"SEMI": THE PROPOSED SEMISUPERVISED TASK LEARNING. "ACTIVE": THE PROPOSED ACTIVE DATA SELECTION STRATEGY

| Dataset | Core-set | LL4AL | VAAL | SRAAL | Base | Base+Semi | Base+Active | Base+Semi+Active |
|---|---|---|---|---|---|---|---|---|
| Cifar-10 | 91.8 | 94.1 | 92.0 | 92.5 | 91.8 | 92.2 (+0.4) | 94.2 (+2.4) | **94.5** (+2.7) |
| Cifar-100 | 65.0 | 65.2 | 65.4 | 66.2 | 62.3 | 66.1 (+3.8) | 67.3 (+5.0) | **68.5** (+6.2) |

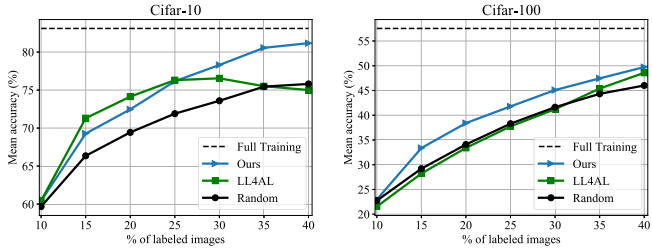

Fig. 9. Noisy active learning results. The labels of training data are randomly perturbed with a probability of 20%.

that more training data can help model training when there are more noisy labels.

## C. Active Learning for Semantic Segmentation

*1) Experimental Setup:* To validate the active learning performance on more complex and large-scale scenarios, we study the semantic segmentation task with the Cityscapes dataset [73] which is a large-scale driving video dataset collected from urban street scenes. Semantic segmentation addresses the pixel-level classification task and its annotation cost is much higher. Following the settings in [17] and [18], we employ the 22-layer dilated residual network (DRN-D-22) [74] as the semantic segmentation model. We report the mean Intersection over Union (mIoU) on the validation set of Cityscapes. We compare our method against SRAAL [18], VAAL [17], QBC [41], Core-set [6], MC-Dropout [28], and the random selection.

*2) Results:* Fig. 10 shows the semantic segmentation performances of different active learning methods on Cityscapes. Our method outperforms the other baselines in terms of mIoU. The results demonstrate the competence of our method on the challenging semantic segmentation task. Note that in the proposed method, neither task model training nor data sampling needs to exploit extra domain knowledge. Therefore, our method is independent of tasks. Moreover, the image size of Cityscapes (i.e., 2048 × 1024) is much larger than that of the classification benchmarks, indicating that our method is not sensitive to the data complexity. These advantages make our method a competitive candidate for complex real-world applications.

## D. Ablation Study

*1) Active Data Sampling Strategy:* Fig. 11 compares different active data sampling strategies on Cifar-10 and Cifar-100. CyclicOD and EMAOD are two variants of TOD, where

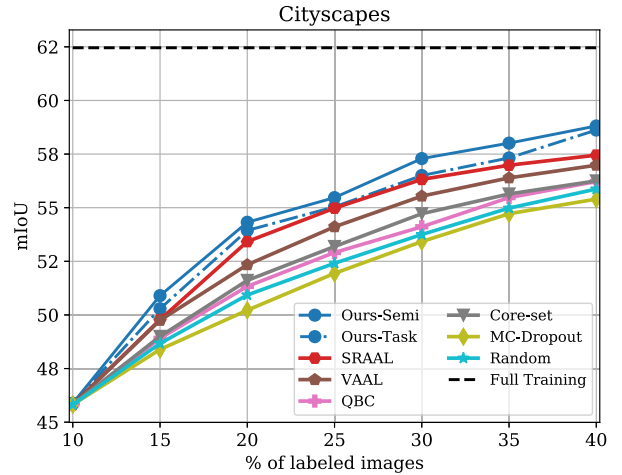

Fig. 10. Active learning results of semantic segmentation on CityScapes dataset.

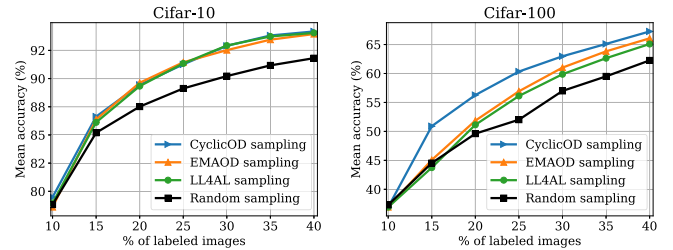

Fig. 11. Ablation on active data sampling.

CyclicOD employs the model at the end of last cycle as the baseline model while EMAOD employs an EMA of the previous models as the baseline model. LL4AL [16] uses a learned loss prediction module to sample the unlabeled data. Fig. 11 shows that the proposed sampling strategies, i.e., EMAOD and CyclicOD, outperform random sampling and LL4AL sampling on both datasets, validating the effectiveness of TOD-based sampling strategy. CyclicOD performs better than EMAOD on Cifar-100, thus we employ COD as the sampling strategy in our active learning approach.

*2) Semisupervised Task Learning:* To evaluate the necessity of semisupervised task learning in active learning, Fig. 12 compares different loss functions on Cifar-10 and Cifar-100. CyclicOD loss and EMAOD loss are two TOD-based unsupervised learning criteria. They are minimized on the unlabeled data, and, the settings of their baseline models are identical to those in the study of sampling strategy as discussed above. LL4AL loss [16] minimizes the distance between the predicted loss and the real task loss, and it needs the data labels. All the auxiliary losses are used in a combination with the task loss. The full pipeline and the training with only the task loss are also included in comparison. We observe that either the EMAOD loss or the CyclicOD loss can help to improve the performance, and either of them shows a larger performance improvement than the LL4AL loss. The EMAOD loss demonstrates a more stable performance than the CyclicOD loss, indicating that directly applying COD to unsupervised training may lead to an unstable model training. A moving average of previous model states enables a more stable unsupervised training.

TABLE II
CLASSWISE PERFORMANCES ON CITYSCAPES, WHERE 40% LABELED DATA ARE USED FOR TRAINING. "PROPORTION" DENOTES THE PROPORTIONS OF CLASSES AT PIXEL LEVEL. "T" IS THE MODEL TRAINED USING ONLY THE TASK LOSS. "T + U" IS THE MODEL TRAINED USING BOTH THE TASK LOSS AND THE PROPOSED TOD-BASED UNSUPERVISED LOSS

| Class ID | 1 | 2 | 3 | 4 | 5 | 6 | 7 | 8 | 9 | 10 | 11 | 12 | 13 | 14 | 15 | 16 | 17 | 18 | 19 | Ave |
|---|---|---|---|---|---|---|---|---|---|---|---|---|---|---|---|---|---|---|---|---|
| **Proportion** (%) | 37.4 | 5.4 | 22.3 | 0.8 | 0.8 | 1.5 | 0.2 | 0.7 | 17.3 | 0.8 | 3.3 | 1.3 | 0.2 | 6.5 | 0.3 | 0.4 | 0.1 | 0.1 | 0.7 | - |
| **T** (mIoU) | 92 | 67 | 82 | 16 | 27 | 53 | 53 | 63 | 87 | 42 | **84** | 71 | 43 | 86 | 19 | 34 | **20** | 31 | 69 | 54.7 |
| **T + U** (mIoU) | **95** | **72** | **87** | **24** | **28** | **56** | **59** | **72** | **90** | **49** | **84** | **77** | **52** | **90** | **24** | **42** | 8 | **39** | **73** | **58.9** |

TABLE III
MODEL-LEVEL MODEL SELECTION RESULTS. WE EVALUATE WHETHER THE BEST MODEL IS IN SCOPE OF THE TOP-$k$ MODELS RETURNED BY EACH METHOD. THE AVERAGE TOP-$k$ PERFORMANCE (%) IS REPORTED. THE NUMBER IN PARENTHESIS IS THE NUMBER OF CANDIDATE MODELS. **BOLD** DENOTES THE BEST PERFORMANCE. <u>UNDERLINE</u> DENOTES THE SECOND-BEST PERFORMANCE

| | **Cifar-10** | | | | **Cifar-100** | | | | **SVHN** | | | |
|---|---|---|---|---|---|---|---|---|---|---|---|---|
| | Top-1(3) | Top-1(5) | Top-1(10) | Top-3(10) | Top-1(3) | Top-1(5) | Top-1(10) | Top-3(10) | Top-1(3) | Top-1(5) | Top-1(10) | Top-3(10) |
| Train loss | **68.7** | **54.4** | 34.3 | 75.2 | <u>62.2</u> | <u>53.3</u> | <u>49.6</u> | <u>78.4</u> | 57.6 | 37.9 | 20.1 | 53.5 |
| Least-C | 59.1 | 48.5 | 38.4 | 72.9 | 55.7 | 42.3 | 33.4 | 57.0 | 61.8 | 52.7 | 41.8 | 78.5 |
| Margin-C | 59.8 | 50.4 | <u>39.7</u> | 74.5 | 56.2 | 43.5 | 36.2 | 61.0 | 61.3 | 52.0 | 44.7 | 84.1 |
| Ratio-C | 61.9 | 48.2 | **41.9** | **77.8** | 49.9 | 41.4 | 32.1 | 53.0 | **71.1** | **62.4** | **55.9** | 87.7 |
| Entropy | 60.4 | 47.1 | 27.4 | 66.3 | 52.2 | 40.1 | 35.3 | 63.3 | 52.2 | 40.9 | 30.0 | 64.5 |
| LL4AL | 39.6 | 22.2 | 9.1 | 30.8 | 30.1 | 14.2 | 4.0 | 17.7 | 38.3 | 25.3 | 18.0 | 28.0 |
| TOD (ours) | <u>63.9</u> | <u>50.8</u> | 36.6 | <u>77.4</u> | **65.8** | **59.3** | **53.9** | **87.4** | <u>63.4</u> | <u>56.6</u> | <u>49.9</u> | **85.2** |

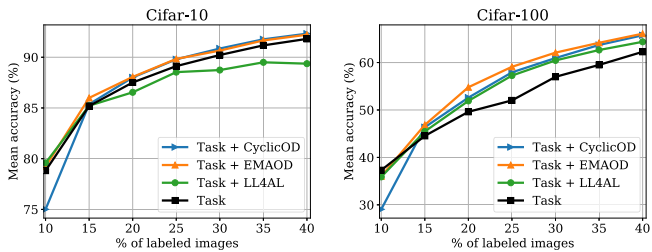

Fig. 12. Ablation on semisupervised task learning.

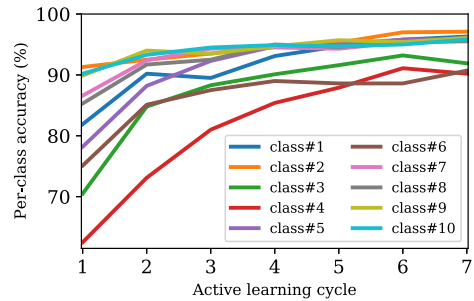

Fig. 13. Per-class accuracy on Cifar-10 using the proposed active learning method.

Table II shows the per-class performance of standard task model training on Cityscapes, where 40% labeled data are observable. The row of "Proportion" in Table II shows the pixel-level proportion of every class, indicating a severe class imbalance problem of Cityscapes. We compare the models trained without (i.e., "T") and with (i.e., "T + U") the unsupervised loss. The semisupervised learning yields better results on 18 out of 19 classes. More importantly, the semisupervised learning shows more significant performance improvements on the minority classes than the majority classes, demonstrating that the unsupervised loss imparts robustness to the task model to handle the class imbalance issue.

*3) Per-Class Performance:* Fig. 13 shows the per-class accuracy on Cifar-10 with the proposed active learning method. The accuracies of the classes are improved along with the increasing of active learning cycles in most cases, such that the performance improvement is not biased toward certain classes. The accuracies of class#3 and class#4 decrease from the sixth cycle to the seventh cycle, mainly due to the overfitting.

*E. Model Selection*

In this section, we study the performance of test data-aware model selection methods on Cifar-10, Cifar-100, and SVHN datasets. The baseline methods consist of the uncertainty estimation methods and the loss estimation methods. The uncertainty estimation methods include least confidence [10], [75], margin of confidence [33], [76], ratio of confidence [32], and entropy-based method [10], [11], [35]. The loss estimation methods include training loss, LL4AL [16], and our TOD method.

We first investigate the model-level model selection performance in Table III. In Table III, we evaluate whether the groundtruth best model is in scope of the top-$k$ models returned by each method, then report the average top-$k$ performance. We observe that our TOD method achieves the best performance on Cifar-100 and the second-best performance on Cifar10 and SVHN. Except TOD, the ratio of confidence method also performs well on three datasets. The results demonstrate the effectiveness of TOD for model selection, as it can pick out the best model from a pool of candidate models by evaluating them on the test set.

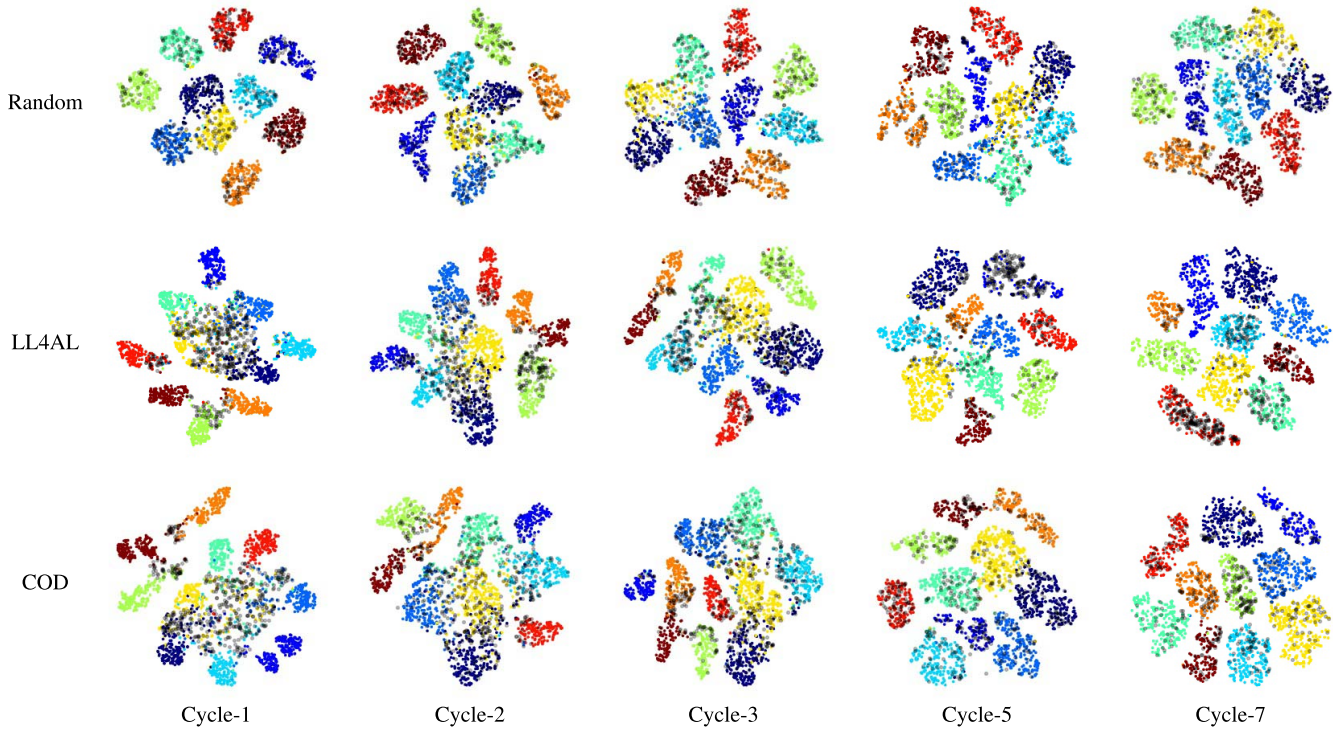

Fig. 14. T-SNE visualization on Fashion-MNIST dataset [65]. **Colored points**: The labeled samples in the training pool. Each color refers to one semantic category. **Gray points**: The samples selected for oracle annotation after an active learning cycle. "Random" denotes random selection.

TABLE IV
FINAL TEST ACCURACY (%) OF SAMPLE-LEVEL MODEL SELECTION FROM A POOL OF TEN CANDIDATE MODELS. *Min/Mean/Max* DENOTE THE LOWEST/AVERAGE/HIGHEST PERFORMANCE OF SINGLE MODELS IN THE POOL

|  | **Cifar-10** | **Cifar-100** | **SVHN** |
|---|---|---|---|
| *min* | 92.41 | 73.29 | 95.20 |
| *mean* | 93.13 | 74.21 | 95.47 |
| *max* | 93.68 | 75.07 | 95.86 |
| Least-C | 95.49 | 79.44 | 96.77 |
| Margin-C | 95.47 | 79.38 | 96.76 |
| Ratio-C | 95.49 | 79.30 | 96.74 |
| Entropy | 95.45 | 79.29 | 96.79 |
| LL4AL | 93.15 | 74.26 | 95.66 |
| TOD (ours) | **95.54** | **79.49** | **96.85** |

TABLE V
ABLATION ON THE NUMBER OF GD STEPS USED IN TOD FOR MODEL SELECTION. THE FINAL TEST ACCURACY (%) OF SAMPLE-LEVEL MODEL SELECTION IS REPORTED

|  | **Cifar-10** | **Cifar-100** | **SVHN** |
|---|---|---|---|
| TOD (1-epoch) | 95.54 | **79.49** | 96.85 |
| TOD (5-iter) | 95.37 | 78.80 | 96.65 |
| TOD (20-iter) | 95.33 | 79.00 | 96.73 |
| TOD (50-iter) | 95.48 | 79.02 | 96.88 |
| TOD (2-epoch) | **95.59** | 78.97 | 96.82 |
| TOD (5-epoch) | 95.49 | 79.18 | **97.02** |

We further study the sample-level model selection performance in Table IV. In this experiment, the top-$k$ metric makes less sense as it needs to be averaged over all the test samples. Thus, we report the average test accuracy in Table IV. Each method evaluates the uncertainty/loss of a single sample, then selects the best model for that sample. The results are averaged over all the test samples. We observe that our TOD method outperforms the baseline methods on all three datasets. TOD also significantly outperforms the best single model, e.g., 95.54% versus 93.68% on Cifar-10 and 79.49% versus 75.07% on Cifar-100. The results further demonstrate the effectiveness of TOD for model selection.

As discussed in Section VI-E, the GD iteration interval between the evaluated and baseline models is an important hyperparameter of TOD-based model selection method. Therefore, we specifically study its effect on model selection in Table V. We have the following observations: 1) different GD iteration intervals may induce a slight performance difference and 2) the best iteration interval is different with respect to the datasets. We use 1 epoch as the default iteration interval for other model selection experiments.

### F. Time Efficiency

This article proposes to use COD for active data sampling. Table VI evaluates the time taken for one active sampling iteration using different active learning methods. On all the three image classification datasets with different number and size of sampling images, COD is faster than the existing active learning methods. COD is task-agnostic and more efficient, since it only relies on the task model itself and it does not introduce extra learnable models such as the adversarial network (VAAL) [17] or the loss prediction module (LL4AL) [16].

TABLE VI
TIME (s) TAKEN FOR ONE ITERATION OF ACTIVE SAMPLING USING AN
NVIDIA GTX 1080TI GPU. *Number of Sampling Images* AND
*Size of Images* ARE SHOWN FOR EACH DATASET

| Method | Cifar-10 2.5K, $32^2$ | SVHN 3.6K, $32^2$ | Caltech-101 0.4K, $224^2$ | Extra model? |
|---|---|---|---|---|
| Coreset [6] | 91.4 | 168.7 | 48.2 | × |
| VAAL [17] | 13.0 | 17.2 | 32.6 | √ |
| LL4AL [16] | 7.7 | 10.8 | 39.6 | √ |
| COD (ours) | **7.2** | **10.1** | **26.9** | × |

*G. T-SNE Visualization*

As shown in Fig. 14, we further conduct t-SNE visualization [77], which allocates every high-dimensional data point a location in a 2-D map, to better understand the active data selection strategies. We study three active selection strategies including LL4AL [16], our COD, and a random selection baseline on Fashion-MNIST dataset [65]. We use the model trained under each strategy to compute the after-softmax class probabilities of all the labeled and unlabeled samples. The probabilities are used as the input data points for t-SNE visualization. In Fig. 14, we use colored points to denote the labeled samples in training pool and gray points to denote the selected samples after each active learning cycle.

We have the following observations from Fig. 14. First, in early active learning cycles, most of the samples selected by COD and LL4AL are on the boundaries of classes, however, the samples selected by the random method are distributed uniformly within the area of each class. Those samples on class boundaries remain more uncertainty with respect to the task models, thus they can significantly help model training in the future. Second, the above observation is more distinct in early cycles than in late cycles, i.e., more selected samples appears in the area of each class in late cycles. The reason is that there are less samples of high uncertainty in late cycles, Third, the t-SNE clustering results of COD are slightly better than those of Random and LL4AL in late cycles. It demonstrates the effectiveness of COD in active data selection.

## VII. CONCLUSION

In this article, we have presented a simple yet effective deep active learning approach. The core of our approach is a measurement TOD which estimates the loss of unlabeled samples by evaluating the discrepancy of outputs given by task models at different gradient descend steps. We have theoretically shown that TOD lower-bounds the accumulated sample loss. On basis of TOD, we have developed an unlabeled data sampling strategy and a semisupervised training scheme for active learning. We have also presented a TOD-based model selection method, which is able to select the models of potentially high testing accuracy from a model pool. Due to the simplicity of TOD, the methods proposed in this article are efficient, flexible, and easy to implement in practice. Extensive experiments have demonstrated the effectiveness of our methods on various image classification and semantic segmentation tasks. In future work, it is a meaningful direction to combine the proposed methods with diversity-aware active learning methods to further reduce the redundancy in selected samples. We plan to incorporate diversity measures into our algorithms, as well as applying our methods to clusters [26] or core-sets [6], [70] of data points.

## APPENDIX

*A. Proofs*

*Theorem 2:* With an appropriate setting of learning rate $\eta$

$$D_t^{\{1\}}(x) \leq \eta\sqrt{2L_t(x)}\|\nabla_w f(x;w_t)\|^2. \quad (11)$$

*Proof:* We apply one-step GD to $w_t$ then using first-order Taylor series

$$\begin{aligned}
D_t^{\{1\}}(x) &\stackrel{\text{def}}{=} \|f(x;w_{t+1}) - f(x;w_t)\| \\
&= \|f(x;w_t - \eta\nabla_{w_t}L_t(x)) - f(x;w_t)\| \\
&= \|f(x;w_t) - \eta\nabla_w f(x;w_t)^T\nabla_w L_t(x) - f(x;w_t)\| \\
&= \|-\eta\nabla_w f(x;w_t)^T\nabla_w L_t(x)\|. \quad (12)
\end{aligned}$$

Recall that

$$\nabla_w L_t(x) = (y - f(x;w_t)) \cdot \nabla_w f(x;w_t). \quad (13)$$

By substituting (13) into (12)

$$\begin{aligned}
D_t^{\{1\}}(x) &= \eta\|(y - f(x;w_t)) \cdot \nabla_w f(x;w_t)^T\nabla_w f(x;w_t)\| \\
&\leq \eta\|(y - f(x;w_t))\| \cdot \|\nabla_w f(x;w_t)\|^2 \\
&= \eta\sqrt{2L_t(x)}\|\nabla_w f(x;w_t)\|^2. \quad (14)
\end{aligned}$$

*Corollary 3:* With an appropriate setting of learning rate $\eta$

$$D_t^{\{T\}}(x) \leq \sqrt{2}\eta \sum_{\tau=t}^{t+T-1}\left(\sqrt{L_\tau(x)}\|\nabla_w f(x;w_\tau)\|^2\right). \quad (15)$$

*Proof:*

$$\begin{aligned}
D_t^{\{T\}}(x) &\stackrel{\text{def}}{=} \|f(x;w_{t+T}) - f(x;w_t)\| \\
&\leq \sum_{\tau=t}^{t+T-1}\|f(x;w_{\tau+1}) - f(x;w_\tau)\| \\
&\leq \sqrt{2}\eta \sum_{\tau=t}^{t+T-1}\left(\sqrt{L_\tau(x)}\|\nabla_w f(x;w_\tau)\|^2\right). \quad (16)
\end{aligned}$$

*Remark 2:* For a linear layer $\phi(x;W)$ with ReLU activation, the Lipschitz constant $\mathcal{L}(W) \leq \|x\|$.

*Proof:*

$$\begin{aligned}
\|\phi(x;W+r) &- \phi(x;W)\| \\
&= \|\max(0,(W+r)^Tx + b) - \max(0,W^Tx + b)\| \\
&\leq \|r^Tx\| \\
&\leq \|x\| \cdot \|r\|. \quad (17)
\end{aligned}$$

Therefore, the Lipschitz constant $\mathcal{L}(W) \leq \|x\|$.

TABLE VII
SUMMARY OF DATASETS USED IN THE EXPERIMENTS. "IMAGE SIZE" IS THE SIZE OF IMAGES USED FOR TRAINING (AFTER PREPROCESSING)

| Dataset | Task | Content | #Classes | Image size | Train | Val | Test |
| --- | --- | --- | --- | --- | --- | --- | --- |
| Cifar-10 | image classification | natural images | 10 | 32×32 | 45,000 | 5,000 | 10,000 |
| Cifar-100 | image classification | natural images | 100 | 32×32 | 45,000 | 5,000 | 10,000 |
| Fashion-MNIST | image classification | fashion images | 10 | 28×28 | 50,000 | 10,000 | 10,000 |
| SVHN | image classification | street view house numbers | 10 | 32×32 | 65,931 | 7,326 | 26,032 |
| Caltech-101 | image classification | natural images | 101 | 224×224 | 7,316 | 915 | 915 |
| STL-10 | image classification | natural images | 10 | 96×96 | 4,500 | 500 | 8,000 |
| Cityscapes | semantic segmentation | driving video frames | 19 | 688×688 | 2,675 | 300 | 500 |

TABLE VIII
SUMMARY OF IMPLEMENTATION DETAILS ON EACH DATASET. "START" IS THE NUMBER OF INITIALLY LABELED SAMPLES AND "BUDGET" IS THE NUMBER OF NEWLY ANNOTATED SAMPLES IN EACH CYCLE. "CYCLE" IS THE NUMBER OF ACTIVE LEARNING CYCLES. $\alpha$ IS THE EMA DECAY RATE AND $\lambda$ IS THE WEIGHT FOR UNSUPERVISED LOSS

| Dataset | Start | Budget | Cycle | Optimizer | LR | Momentum | Decay | Epochs | Batch | $\alpha$ | $\lambda$ |
| --- | --- | --- | --- | --- | --- | --- | --- | --- | --- | --- | --- |
| Cifar-10 | 10% | 5% | 7 | SGD | 0.1 | 0.9 | $5\times10^{-4}$ | 200 | 128 | 0.999 | 0.05 |
| Cifar-100 | 10% | 5% | 7 | SGD | 0.1 | 0.9 | $5\times10^{-4}$ | 200 | 128 | 0.999 | 0.05 |
| Fashion-MNIST | 10% | 5% | 7 | SGD | 0.1 | 0.9 | $5\times10^{-4}$ | 200 | 128 | 0.999 | 0.05 |
| SVHN | 10% | 5% | 7 | SGD | 0.1 | 0.9 | $5\times10^{-4}$ | 200 | 128 | 0.999 | 0.05 |
| Caltech-101 | 10% | 5% | 7 | SGD | 0.01 | 0.9 | $5\times10^{-4}$ | 50 | 64 | 0.999 | 0.05 |
| STL-10 | 10% | 5% | 7 | SGD | 0.1 | 0.9 | $5\times10^{-4}$ | 200 | 64 | 0.999 | 0.05 |
| Cityscapes | 10% | 5% | 7 | Adam | $5\times10^{-4}$ | - | - | 40 | 4 | 0.999 | 0.05 |

*Corollary 4:* With appropriate settings of a learning rate $\eta$ and a constant $C$

$$D_t^{\{T\}}(x) \leq \sqrt{2T}\eta C \sqrt{\sum_{\tau=t}^{t+T-1} L_\tau(x)}. \qquad (18)$$

*Proof:* By substituting $\|\nabla_w f\|^2 \leq C$ into Corollary 3 then applying Cauchy–Schwarz inequality, we have

$$\begin{aligned} D_t^{\{T\}}(x) &\leq \sqrt{2}\eta C \sum_{\tau=t}^{t+T-1} \sqrt{L_\tau(x)} \\ &\leq \sqrt{2T}\eta C \sqrt{\sum_{\tau=t}^{t+T-1} L_\tau(x)}. \end{aligned} \qquad (19)$$

### B. Experimental Details of Image Classification

*1) Datasets:* We evaluate the active learning methods on four common image classification datasets, including Cifar-10 [64], Cifar-100 [64], Fashion-MNIST [65], SVHN [66], Caltech-101 [67], and STL-10 [68]. CIFAR-10 and CIFAR-100 consist of 50 000 training images and 10 000 testing images with the size of 32 × 32. CIFAR-10 has ten categories and CIFAR-100 has 100 categories. SVHN consists of 73 257 training images and 26 032 testing images with the size of 32 × 32. SVHN has ten classes of digit numbers from "0" to "9." Fashion-MNIST consists of 60 000 training images and 10 000 testing images with the size of 28 × 28. Fashion-MNIST has ten classes of fashion images including t-shirt, trouser, dress, coat, and bag. For training on Cifar-10, Cifar-100, Fashion-MNIST, and SVHN, we randomly crop 32 × 32 images from the 36 × 36 zero-padded images. Caltech-101 consists of 9146 images with the size of 300 × 200. Caltech-101 has 101 semantic categories as well as a background category that there are about 40–800 images per category. By following [18], we use 90% of the images for training and 10% of the images for testing. On Caltech-101, we resize the images to 256 × 256 and crop 224 × 224 images at the center. STL-10 consists of 5000 training images and 8000 testing images with the size of 96 × 96, serving as a high-resolution extension of Cifar-10. On STL-10, we resize the images to 128 × 128 and crop 96 × 96 images at the center. Random horizontal flip and normalization are applied to all the image classification datasets. We summarize the details of the datasets in Table VII.

*2) Implementation Details:* We employ ResNet-18 [69] as the image classification model. On all the image classification datasets, the labeling ratio of each active learning cycle is 10%, 15%, 20%, 25%, 30%, 35%, and 40%, respectively. In an cycle, The model is learned for 200 epochs using an SGD optimizer with a learning rate of 0.1, a momentum of 0.9, a weight decay of $5 \times 10^{-4}$, and a batch size of 128. After 80% of the training epochs, the learning rate is decreased to 0.01. We summarize the implementation details in Table VIII.

### C. Experimental Details of Semantic Segmentation

*1) Dataset:* We evaluate the active learning methods for semantic segmentation on the Cityscapes dataset [73]. Cityscapes is a large-scale driving video dataset collected from urban street scenes of 50 cities. It consists of 2975 training images and 500 testing images with the size of 2048 × 1024. By following [17], we convert the dataset from the original 30 classes into 19 classes. We crop 688 × 688 images from the original images for training. Random horizontal flip and normalization are applied to the images.

*2) Implementation Details:* We employ the DRN-D-22 [74] as the semantic segmentation model. The labeling ratio of each

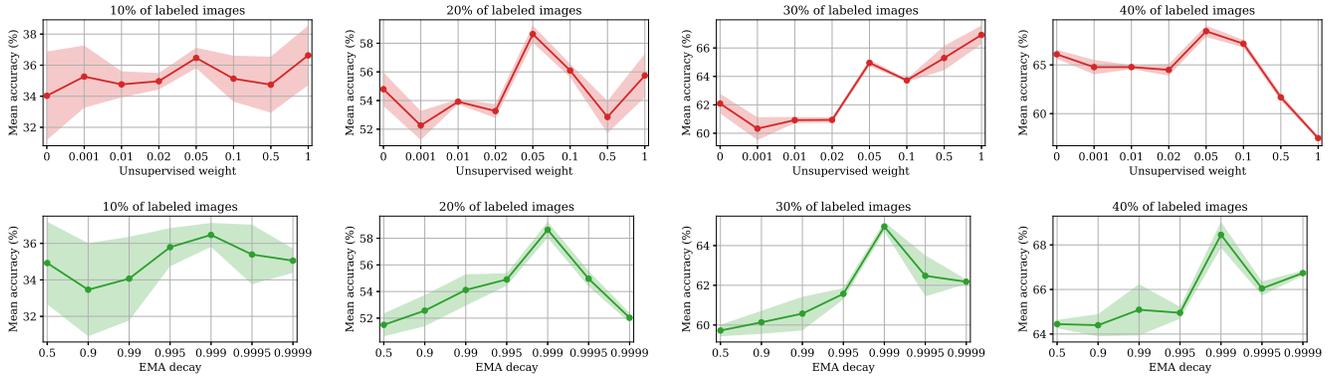

Fig. 15. Empirical study on unsupervised loss weight $\lambda$ and EMA decay rate $\alpha$. The study is conducted on Cifar-100 with 10%, 20%, 30%, and 40% of labeled images, respectively. $\lambda = 0.05$ and $\alpha = 0.999$ achieve the best performances.

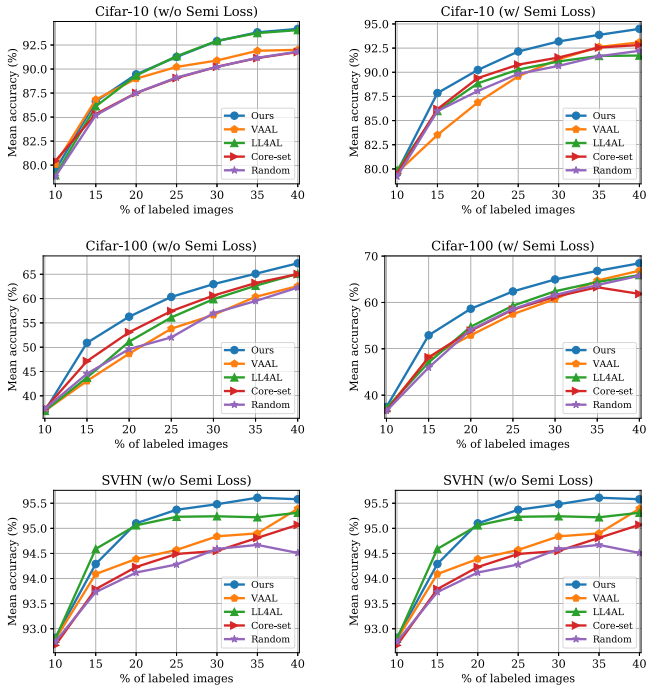

Fig. 16. Performance of benchmark active learning methods trained *without* (left) and *with* (right) the unsupervised loss on three datasets.

active learning cycle is 10%, 15%, 20%, 25%, 30%, 35%, and 40%, respectively. In an cycle, the model is learned for 40 epochs using an Adam optimizer [78] with a learning rate of $5 \times 10^{-4}$ and a batch size of 4.

### D. Study on Hyperparameters

The unsupervised learning plays a vital role in training the task model in active learning. Here we further investigate the hyperparameter selection for our proposed unsupervised learning method. The hyperparameters include the unsupervised loss weight $\lambda$ and the EMA decay rate $\alpha$. We conduct empirical studies using our full active learning pipeline on Cifar-100 dataset to investigate the performance variation with different $\lambda$ and $\alpha$. Fig. 15 shows the results on labeling budgets of 10%, 20%, 30%, 40%, respectively. In most of the cases, $\lambda = 0.05$ and $\alpha = 0.999$ achieve the best performance. Therefore, we adopt $\lambda = 0.05$ and $\alpha = 0.999$ for all the experiments in this article, wherever EMA is involved.

### E. More Results of Active Data Selection Strategy

We additionally compare the active data selection strategies by training the task model with and without the unsupervised loss, respectively. Fig. 16 shows that our method achieves superior performances on most of the datasets and settings (either with or without unsupervised loss), demonstrating its effectiveness in active data selection.

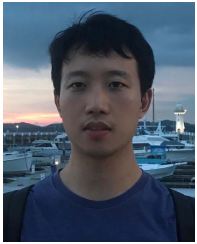
**Siyu Huang** received the B.E. and Ph.D. degrees in information and communication engineering from Zhejiang University, Hangzhou, China, in 2014 and 2019, respectively.

He was a Visiting Scholar with the School of Computer Science, Language Technologies Institute, Carnegie Mellon University, Pittsburgh, PA, USA, in 2018; a Research Scientist with the Big Data Laboratory, Baidu Research, Beijing, China, from 2019 to 2021; and a Research Fellow with the School of Electrical and Electronic Engineering, Nanyang Technological University, Singapore, in 2021. He is currently a Post-Doctoral Fellow with the John A. Paulson School of Engineering and Applied Sciences, Harvard University, Cambridge, MA, USA. He has published more than 20 papers in top-tier computer science journals and conferences. His research interests include computer vision, multimedia analysis, and deep learning.

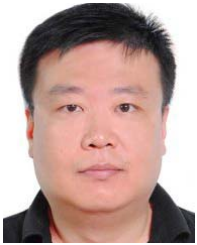
**Tianyang Wang** received the B.S. and M.S. degrees from Jilin University, Changchun, China, in 2006 and 2010, respectively, and the Ph.D. degree in computer science from Southern Illinois University, Carbondale, IL, USA, in 2015.

He was a Visiting Researcher with Baidu Research, Beijing, China, in 2019. He is currently an Assistant Professor with the Department of Computer Science and Information Technology, Austin Peay State University, Clarksville, TN, USA. His research interests include artificial intelligence, machine learning, and computer vision.

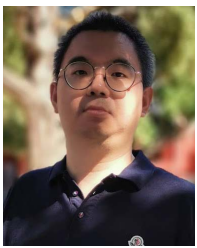
**Haoyi Xiong** (Senior Member, IEEE) received the Ph.D. degree in computer science from the Telecom SudParis, University of Paris VI, Paris, France, in 2015.

From 2015 to 2016, he was a Post-Doctoral Research Associate with the Department of Systems and Information Engineering, University of Virginia, Charlottesville, VA, USA. From 2016 to 2018, he was an Assistant Professor with the Department of Computer Science, Missouri University of Science and Technology, Rolla, MO, USA (formerly known as University of Missouri at Rolla). He is currently with the Big Data Laboratory, Baidu Research, Beijing, China. He has published more than 60 papers in top computer science conferences and journals, such as the ACM International Conference on Pervasive and Ubiquitous Computing (UbiComp), International Conference on Learning Representations (ICLR), RTSS, Association for the Advancement of Artificial Intelligence (AAAI), International Joint Conference on Artificial Intelligence (IJCAI), IEEE TRANSACTIONS ON MOBILE COMPUTING, IEEE TRANSACTIONS ON NEURAL NETWORKS AND LEARNING SYSTEMS, and *ACM Transactions on Knowledge Discovery from Databases*. His current research interests include automated deep learning, ubiquitous computing, artificial intelligence, and cloud computing.

Dr. Xiong was a recipient of the Best Paper Award from the IEEE International Conference on Ubiquitous Intelligence and Computing (UIC) 2012 and the Outstanding Ph.D. Thesis Runner Up Award from CNRS SAMOVAR 2015.

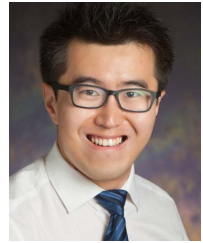
**Bihan Wen** (Member, IEEE) received the B.Eng. degree in electrical and electronic engineering from Nanyang Technological University, Singapore, in 2012, and the M.S. and Ph.D. degrees in electrical and computer engineering from the University of Illinois at Urbana–Champaign, Champaign, IL, USA, in 2015 and 2018, respectively.

He is currently a Nanyang Assistant Professor with the School of Electrical and Electronic Engineering, Nanyang Technological University, Singapore. His research interests include machine learning, computational imaging, computer vision, and image and video processing.

Dr. Wen was a recipient of the 2016 Yee Fellowship and the 2012 Professional Engineers Board Gold Medal. He was also a recipient of the Best Paper Runner Up Award from the IEEE International Conference on Multimedia and Expo in 2020. He has been an Associate Editor of IEEE TRANSACTIONS ON CIRCUITS AND SYSTEMS FOR VIDEO TECHNOLOGY since 2022.

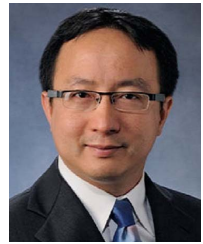
**Jun (Luke) Huan** (Senior Member, IEEE) received the B.S. degree from Peking University, Beijing, China, in 1997, the M.S. degree from Oklahoma State University, Stillwater, OK, USA, in 2000, and the Ph.D. in computer science from the University of North Carolina, Chapel Hill, NC, USA, in 2006.

He served as a Distinguished Scientist and the Head of the Big Data Laboratory, Baidu Research, Beijing, China, and the Chief Scientist and the Chief Executive Officer of Styling AI. From 2006 to 2018, he served as the Charles and Mary Jane Spahr Professor at the Department of Electrical Engineering and Computer Science, The University of Kansas, Lawrence, KS, USA. From 2015 to 2018, he also served as the Program Director for the U.S. National Science Foundation, in charge of the Big Data Program of NSF. He is currently a Principal Applied Scientist at the AWS AI Lab, USA. He has authored more than 100 peer-reviewed papers in leading conferences and journals and has graduated ten Ph.D. students. His research interests include data science, AI, deep learning, and data mining.

Dr. Huan was a recipient of the U.S. National Science Foundation Faculty Early Career Development Award in 2009. His group received several best paper awards from leading international conferences, including IEEE International Conference on Data Mining. His service record includes the Program Co-Chair for the IEEE International Conference on Bioinformatics and Biomedicine (BIBM) in 2015 and IEEE Big Data in 2019.

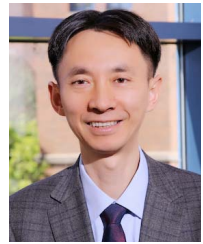
**Dejing Dou** (Senior Member, IEEE) received the B.E. degree in electronic engineering from Tsinghua University, Beijing, China, in 1996, and the Ph.D. degree in artificial intelligence from Yale University, New Haven, CT, USA, in 2004.

He is currently a Professor with the Computer and Information Science Department, University of Oregon, Eugene, OR, USA, where he leads the Advanced Integration and Mining Lab (AIM Lab). He is also the Director of the NSF IUCRC Center for Big Learning (CBL), Alexandria, VA, USA. He is currently on sabbatical leave at Baidu Research, Beijing, from 2019 to 2020. His research interests include ontologies, data mining, data integration, information extraction, and health informatics.